# A Deep Reinforcement Learning (DRL)-Based Transformer Method for Solving the Open Shop Scheduling Problem*

Faezeh Ardali[1†], Mwembezi A. Nyelele[2], and Gerald M. Knapp[1]

[1] *Louisiana State University, Baton Rouge, LA, USA*
[2] *University of Minnesota Duluth, Duluth, MN, USA*

**Abstract**

The open shop scheduling problem (OSSP) arises in many industrial and service settings but remains computationally challenging as the number of jobs and machines increases. While exact methods quickly become intractable, classical dispatching rules and metaheuristics may require substantial tuning to maintain solution quality at large scales. This study develops a Transformer-based scheduling policy for OSSP using an encoder–decoder architecture with multi-head attention. The model is trained on Taillard benchmark instances (4×4, 5×5, 7×7, and 10×10) using only the processing-time matrix as input and produces feasible schedules with makespans typically within 15–30% of best-known values. To evaluate scalability, the trained policy is applied without retraining to randomly generated instances from 40×40 to 100×100 and compared against classical dispatching heuristics, including SPT, LPT, MWKR, and EST. Across these large instances, the Transformer achieved average gaps of 12.89–15.12% relative to a standard lower bound. Compared with EST, the Transformer remained competitive (typically within a modest margin) while substantially outperforming SPT and LPT. These results indicate that a Transformer policy trained on small OSSP instances can generalize to substantially larger problems and provide a feature-light, learning-based alternative to classical dispatching rules.

## 1. Introduction

Scheduling plays a vital role in many industries, including manufacturing, transportation, and computing. It determines how limited resources are allocated to competing tasks over time, directly influencing productivity, cost, and overall system efficiency. Among various scheduling problems, the Open Shop Scheduling Problem (OSSP) is one of the most flexible and general. In OSSP, each job consists of multiple operations that must be processed on different machines, but unlike flow shop or job shop problems, the order of operations is not predetermined. However, the optimization of OSSP remains a major challenge because it is classified as an NP-hard problem [1]. Finding an optimal schedule requires exploring a large solution space that grows exponentially with the number of jobs and machines. As a result, many researchers have focused on developing efficient optimization methods—from classical heuristics to modern artificial intelligence and meta-heuristic algorithms—to find high-quality solutions within reasonable computational time.

Research on OSSP has progressed from exact optimization to data-driven and learning-guided search. Early studies employed mixed-integer programming, branch-and-bound, and constraint programming to obtain optimal schedules, which perform well on small instances but scale poorly [2]. As the problem size grows, heuristic approaches generate feasible solutions more efficiently [3]. Metaheuristics—including simulated annealing, tabu search, genetic algorithms, greedy randomized adaptive search procedure (GRASP), and ant colony optimization—further balance exploration and exploitation, achieving competitive results on standard benchmarks [4]. Hybrid frameworks combining these metaheuristics with exact techniques (e.g., large neighborhood search or fix-and-optimize) improve solution quality and robustness. However, such methods still rely heavily on handcrafted rules and parameter tuning [5]. Consequently, recent studies have introduced machine learning approaches that learn dispatching priorities or search policies (via supervised or reinforcement learning), enhancing performance on large-scale OSSP instances.

---

** This is the author's preprint version of a paper accepted for publication in the IISE Annual Conference & Expo 2026 Proceedings.

† Corresponding author: fardal1@lsu.edu

Recent work applies machine learning to OSSP by learning how to choose the next operation or where to search in the solution space. Supervised models can learn dispatching priorities from historical or simulated schedules, producing fast decisions but with limited ability to anticipate long-term scheduling consequences. Reinforcement learning optimizes policies directly from makespan-based rewards, often using feasibility masks to respect machine and precedence constraints [6]; this can yield strong solutions that are feasible at all times and improve as more computation is allowed but requires careful training and reward design. Beyond shop scheduling, Soleymani et al. apply a double deep Q-learning framework which uses two networks—one to choose the best action and another to evaluate it—reducing overestimation bias to autonomous resource allocation in multi-project construction portfolios, showing that a single deep reinforcement learning (DRL) policy can adapt to varying resource and timing constraints without retraining [7]. Similarly, Amani et al. develop an event-driven deep reinforcement learning dispatcher for post-storm power distribution restoration, where decisions are updated in real time as new information becomes available. Their study shows that actor–critic DRL with feasibility masking can produce fast, adaptive decisions and improve operational performance relative to optimization-based and heuristic baselines [8]. Reinforcement learning has also been explored in other dynamic engineering applications. Chahardoli et al. developed a PPO-based RL framework for HVAC control that uses environmental and physiological data to adapt decisions in real time, improving energy efficiency while maintaining occupant comfort [9]. A complementary line of work uses machine learning to guide, rather than replace, classical search methods—for example, by learning destroy/repair strategies in large-neighborhood search or rollout and value functions in tree search—combining the efficiency of ML with the robustness of traditional optimization [10]. Hybrids that warm-start constraint programming (CP) or mixed-integer programming (MIP) solvers or that predict promising neighborhoods further improve solution quality and runtime on standard benchmarks [11]. Beyond shop scheduling, transformer–DRL architectures have also been used for large-scale sequential decision problems such as post-disaster crew dispatch in power systems, where a transformer policy learns near–real-time restoration sequences while respecting network constraints [12]. Building on these advances, recent attention-based models such as Transformers have shown strong results across combinatorial optimization and in structured shop-scheduling variants such as job-shop scheduling, where operation order is fixed. However, their application to OSSP remains very limited, as the lack of a predetermined operation order introduces additional challenges for learning feasible and efficient sequences, motivating the development of a transformer framework tailored to open shop structure.

In this study, we propose a fully Transformer-based architecture integrated with DRL for the OSSP. This work extends previous approaches that relied on graph-based or single-head attention mechanisms by employing the complete Transformer framework with multi-head attention in both the encoder and decoder [13]. Despite using only processing time as input, the model achieves strong performance on benchmark datasets and generalizes well to larger, randomly generated instances without retraining.

## 2. Methodology

In this research, we employ a reinforcement learning framework to optimize scheduling performance, where the transformer model functions as the actor and a critic network evaluates its performance. The Transformer model encompasses two main components: an encoder and a decoder. Each OSSP instance is represented as a set of $N=J\times M$ job–machine operation nodes (one node per operation). For each node, the input token is constructed from the processing time $p_{j,m}$, and its associated indices (job ID and machine ID). These inputs are first converted into fixed-length numerical representations that can be processed by the Transformer. The encoder is responsible for capturing the underlying features and learning contextual relationships among all nodes through multi-head self-attention and feed-forward layers. Through successive attention and normalization layers, the encoder refines these embeddings by allowing each node to attend to others, thereby incorporating inter-node dependencies and global structural information. The resulting node embeddings then serve as the input to the decoder for generating optimized scheduling decisions.

The decoder acts as the decision-making component of the Transformer framework, translating the contextual information learned by the encoder into an actionable scheduling sequence. At each step, it interacts with the encoded graph to decide which machine should operate next and which job–operation node should be assigned to it. These decisions are driven by multiple internal components that mirror the structure of a standard Transformer decoder. The masked multi-head attention mechanism enables the model to focus on the current scheduling state while preventing the reuse of already completed operations, ensuring temporal consistency. The encoder–decoder attention layer allows the decoder to consult the encoder's outputs (node and graph embeddings), integrating global contextual information from the entire problem instance when selecting the next action. Following this, the feed-forward and normalization layers refine and stabilize the attention outputs, enhancing the reliability of the learned decision representations. Finally, the linear and SoftMax layers transform these refined embeddings into probability distributions over feasible machines or nodes, from which the next scheduling decision is sampled. Through continuous updates to the environment and reapplication of these components, the decoder iteratively constructs a valid schedule that balances both local feasibility and global optimization, ultimately evaluating its performance based on the resulting makespan.

The Transformer section functions as the actor network within the Proximal Policy Optimization (PPO) algorithm [14], while a separate critic network serves as the evaluator by estimating the expected reward. PPO is a reinforcement learning algorithm that follows the actor–critic paradigm, where the actor proposes actions, and the critic assesses their quality to guide the learning process. In this setup, the Transformer actor generates scheduling actions—selecting which machine and operation to process next—based on the learned policy. During training, the Transformer policy interacts with the scheduling environment to generate trajectories defined as ordered sequences of scheduling states, selected actions, and resulting reward tuples from the initial empty schedule ($s_t$, $a_t$, $r_t$) until a complete schedule is constructed. These trajectories are collected by the PPO agent and used to compute policy and value updates. After each rollout (once a complete schedule is constructed), the environment returns a reward defined as the negative makespan, indicating the quality of the generated schedule. The critic network compares the observed reward with its predicted value to calculate the advantage, which quantifies how much better or worse the taken action was compared to the expected outcome. This feedback is used to update the actor through PPO's clipped objective function, ensuring stable and efficient policy improvement. Through this iterative interaction among the Transformer actor, the critic evaluator, and the environment, the model progressively refines its scheduling strategy toward generating near-optimal and efficient solutions. Figure 1 shows the architecture of the proposed Transformer-based DRL model.

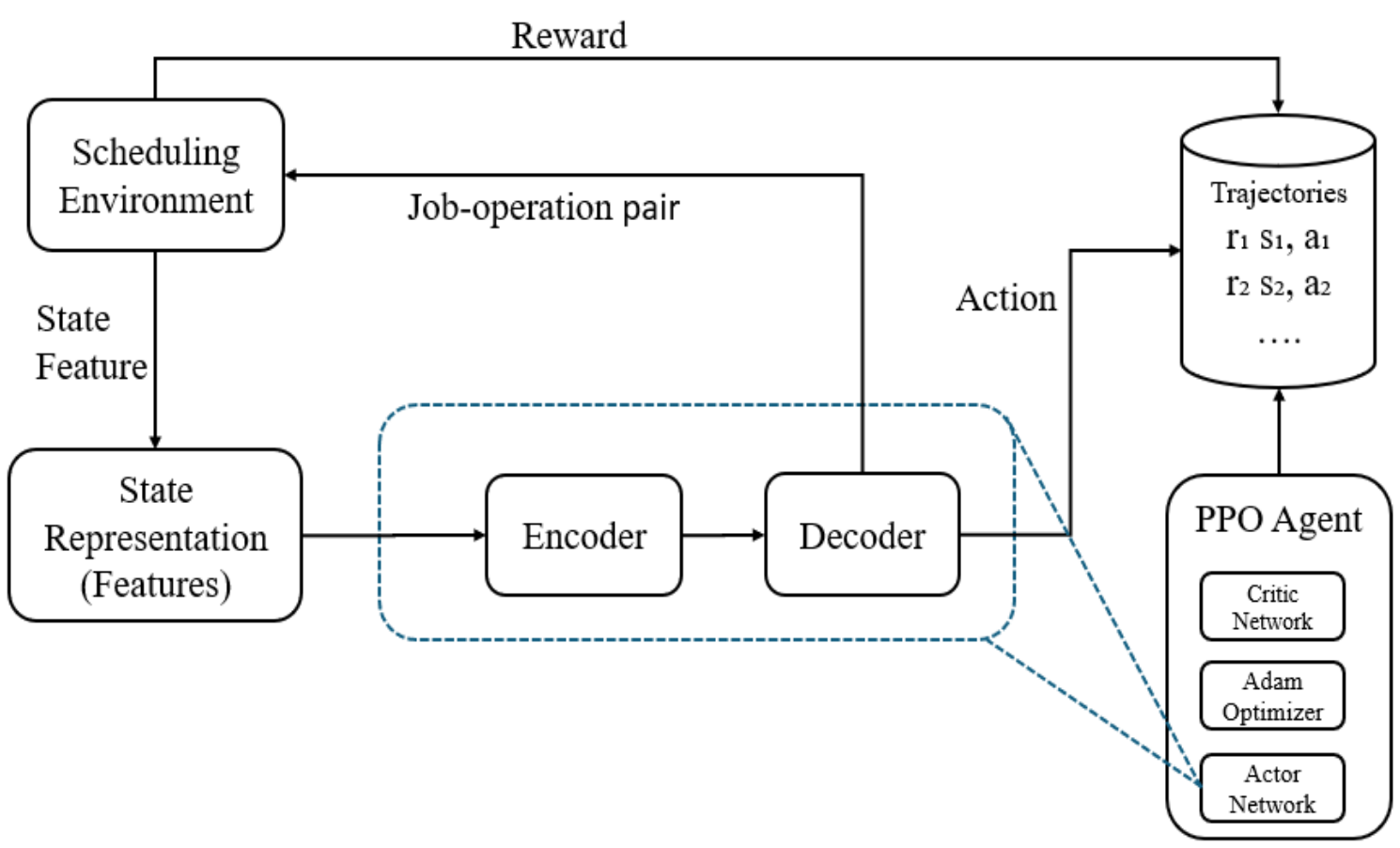

Figure 1. Architecture of the proposed Transformer–PPO framework for OSSP scheduling

## 3. Experiments

### 3.1. Hyperparameter Optimization

Before conducting the main training experiments, a dedicated hyperparameter optimization phase was carried out using Optuna, an automated search framework designed to efficiently explore large parameter spaces. The objective was to identify the most effective configuration of the Transformer–PPO model while balancing model complexity, convergence stability, and computational efficiency. The tuning process systematically varied key Transformer architectural parameters, including the embedding dimension, number of attention heads, number of encoder layers, and tanh clipping, as well as PPO-related parameters such as the learning rate and warm-up coefficient (β). Each parameter was sampled from the predefined ranges summarized in Table 1. Each trial was trained on smaller synthetic OSSP instances (4×4) generated with Taillard's instance generator to enable fast validation cycles [15]. The average makespan across validation samples was used as the performance metric.

Table 1. Search space for hyperparameter optimization using Optuna

| **Parameter** | **Search Range/Values** |
|---|---|
| embed_dim | {32, 64, 128} |
| n_heads | {1, 2, 4, 8} |
| n_encode_layers | {1, 2, 3, 4} |
| tanh_clipping | {5, 10, 20} |
| warmup_beta | {0.6, 0.8, 0.9} |
| lr | $[8\times10^{-5}, 3\times10^{-4}]$ |

The optimization study was executed over 24 trials using Optuna's median pruning strategy to discard underperforming configurations early. The best-performing setting achieved a strong balance between performance and efficiency, yielding the following optimal configuration: embedding dimension = 32, learning rate = $1.15 \times 10^{-4}$, number of encoder layers = 1, number of attention heads = 2, tanh clipping = 20, and warm-up $\beta = 0.8$.

### 3.2. Training Process for DRL

Using the optimized hyperparameters identified in Section 3.1, the Transformer-based DRL model was trained and evaluated on 128000 training size for each of the synthetic OSSP instances of sizes 4 × 4, 5 × 5, 7 × 7, and 10 × 10 (jobs × machines). We used the Adam optimizer with 1,000 rollout samples, a batch size of 512, and 250 batch steps for training each size. Model performance was then evaluated on a separate held-out test set of 10 independently generated instances per size, and all reported performance metrics are averaged over these test instances. Performance was compared with the Taillard benchmark datasets to assess solution quality under controlled conditions.

All training experiments were conducted on a workstation located in the LSU Mechanical and Industrial Engineering Laboratory, equipped with an Intel® Core™ i7-12700K CPU @ 3.60 GHz (12 cores), 32 GB RAM, and an NVIDIA GeForce RTX 3070 Ti GPU (8 GB memory), running Microsoft Windows 10 Enterprise (Build 19045).

OSSP instances were generated using Taillard's dataset-generation Pascal code, with processing times sampled from a discrete uniform distribution U (1,100). Pseudocode for this instance-generation procedure is shown in Figure 3.

**Algorithm 1** Taillard Open Shop Instance Generation

```
1:  procedure GenerateOSSP(n_jobs, n_machines, time_seed, machine_seed)
2:      Initialize matrices d (processing times) and M (machine assignments)
3:      Generate processing times:
4:      for j = 1 to n_jobs do
5:          for i = 1 to n_machines do
6:              d[i, j] ← UniformRandom(1, 99)
7:          end for
8:      end for
9:      Initialize machine assignments:
10:     for j = 1 to n_jobs do
11:         for i = 1 to n_machines do
12:             M[i, j] ← i
13:         end for
14:     end for
15:     Random permutation of machines:
16:     for j = 1 to n_jobs do
17:         for i = 1 to n_machines do
18:             k ← UniformRandom(i, n_machines)
19:             swap(M[i, j], M[k, j])
20:         end for
21:     end for
22:     return d, M
23: end procedure
```

Figure 2. Pseudocode for generating a random Open Shop Scheduling Problem

### 3.3. Generalization capability of the trained model

To evaluate the scalability and generalization capability of the trained model, additional experiments were performed on large-scale OSSP instances. Table 3 summarizes the datasets and their corresponding problem sizes. For each size, 10 random instances were generated for testing, consistent with the setup used for the smaller Taillard-based problems

Table 2. Large-scale experimental configurations

| Type of dataset | Number of jobs | Number of machines | Number of operations | Number of instances |
|---|---|---|---|---|
| 40×40 | 40 | 40 | 1600 | 10 |
| 50×50 | 50 | 50 | 2500 | 10 |
| 70×70 | 70 | 70 | 4900 | 10 |
| 100×100 | 100 | 100 | 10000 | 10 |

In this phase, the trained Transformer-PPO model was directly applied to unseen large-scale instances without additional fine-tuning. Performance was reported in terms of (i) average makespan and (ii) an optimality-gap proxy computed using the standard OSSP workload lower bound [16], defined as

$$LB = \max\left(\max_{j} \sum_{m} p_{j,m}\,,\ \max_{m} \sum_{j} p_{j,m}\right) \tag{1}$$

and the corresponding relative gap

$$Gap(\%) = 100.\frac{C_{max} - LB}{LB} \tag{2}$$

where $p_{j,m}$ denotes the processing time of job j on machine m; j∈{1,…, J} and m∈{1,…, M} with J jobs and M machines. The makespan $C_{max}$ is the completion time of the final operation in the schedule (i.e., the maximum completion time over all jobs and machines). The lower bound LB is the maximum of the largest job workload and the largest machine workload. Gap (%) is the relative gap between the achieved makespan $C_{max}$ and LB, computed by Eq. (2). In addition to reporting absolute makespans, the learned policy was evaluated relative to classical dispatching heuristics: shortest processing time (SPT), longest processing time (LPT), most work remaining (MWKR), and earliest completion time (EST). Solution quality was summarized using the average percentage gap relative to the workload lower bound. Overall, this experiment demonstrates the model's ability to generalize beyond its training distribution and maintain stable solution quality on significantly larger problem sizes.

## 4. Results

This study demonstrates that a Transformer–PPO scheduling policy can (i) consistently produce feasible OSSP schedules, (ii) achieve competitive solution quality on Taillard benchmark instances using only the processing-time matrix as input, and (iii) generalize to substantially larger problem sizes (up to 100×100) without retraining while maintaining stable performance on large random instances relative to a standard workload lower bound.

On Taillard benchmark instances (4×4–20×20), the proposed model produces feasible schedules for all tested sizes and attains makespans within approximately 15–30% of the Taillard reference values on average. The gap to Taillard reference makespans is larger for smaller instances (e.g., 28.80% for 7×7 and 25.88% for 5×5), but decreases as the problem size increases, improving from 22.56% at 10×10 to 18.04% at 15×15 and 15.59% at 20×20. This trend suggests that the learned policy benefits from a richer structure in larger instances and captures scheduling patterns that become more effective as instance complexity grows. Table 4 summarizes these Taillard benchmark results.

Table 3. Results for OSSP Taillard Benchmark Instances

| Size | Avg Taillard | Avg Transformer | Avg Gap (%) |
|---|---|---|---|

| 4×4 | 232.7 | 273.1 | 17.36 |
|---|---|---|---|
| 5×5 | 312.50 | 393.4 | 25.88 |
| 7×7 | 443.3 | 571 | 28.80 |
| 10×10 | 608.8 | 746.2 | 22.56 |
| 15×15 | 912.4 | 1077 | 18.04 |
| 20×20 | 1235.4 | 1428.1 | 15.59 |

To evaluate out-of-distribution generalization, the trained model was applied directly (no fine-tuning) to randomly generated large-scale OSSP instances from 40×40 to 100×100. Since optimal solutions are not available for these large random instances, performance was assessed using two complementary views: (i) comparison against classical dispatching heuristics (SPT, LPT, MWKR, and EST), and (ii) a relative gap computed with respect to a standard workload lower bound, which provides a conservative proxy for solution optimality.

Across large-scale sizes, the Transformer policy performance is similar to the strongest heuristic baseline EST and substantially outperforms simple rules such as SPT and LPT. More importantly, the model maintains stable quality as size grows: the average gap to the lower bound stays within a narrow band of 12.89%–15.12% from 40×40 through 100×100 (12.89% at 40×40, 14.43% at 50×50, 15.12% at 70×70, and 14.74% at 100×100). This consistency supports the claim that a policy trained on smaller instances can generalize effectively to substantially larger OSSP environments under a feature-light input representation. Table 5 reports the large-scale results in detail, including the workload lower bound, heuristic baselines, the Transformer average makespan, and the resulting average gap values.

Table 4. Generalization Results for Large-Scale OSSP Instances

| **Size** | **LB** | **SPT Avg** | **LPT Avg** | **MWKR Avg** | **EST Avg** | **Transformer Avg** | **Avg Gap to LB**(%) |
|---|---|---|---|---|---|---|---|
| 40×40 | 2435 | 6541 | 5158 | 2892 | 2692 | **2646** | 12.89 |
| 50×50 | 2999 | 8479 | 6517 | 3535 | **3252** | 3429 | 14.43 |
| 70×70 | 4047 | 12242 | 8501 | 4710 | **4418** | 4774 | 15.12 |
| 100×100 | 5787 | 17817 | 11397 | 6641 | **6213** | 6639 | 14.74 |

## 5. Conclusion

This study presents a Transformer-based DRL framework for the OSSP. By integrating multi-head attention with Proximal Policy Optimization, the proposed model learns feasible scheduling policies using only the processing-time matrix, avoiding handcrafted features and extensive domain-specific rule design. On Taillard benchmark instances from 4×4 up to 20×20, the model consistently produced valid schedules with moderate optimality gaps. While performance is weaker on very small instances, the gap decreases as problem size increases, improving to 15.59% at 20×20, suggesting that the attention-based policy benefits from richer structure in larger instances. We also evaluated generalization beyond the training regime by applying the trained model—without retraining—to random large-scale OSSP instances from 40×40 to 100×100. Compared with classical dispatching heuristics (SPT, LPT, MWKR, and EST), the learned policy remained within a moderate margin of EST and clearly outperformed simple rules such as SPT and LPT. Since optimal solutions are unavailable for these large instances, quality was additionally assessed using a standard lower bound; the average gap remained stable across scales, ranging from 12.89% to 15.12%.

At the same time, the experiments highlight opportunities for improvement. Full self-attention may limit scalability to very large instances, and the model still shows gaps relative to Taillard reference values. Future work will focus on more efficient attention mechanisms, richer state representations capturing machine/job congestion, and stronger hybridization with classical optimization.

Overall, this work provides evidence that attention-based reinforcement learning is a practical approach for complex shop scheduling. Even with a feature-light input, the proposed policy can be trained on smaller instances and transferred to substantially larger problems without retraining. Further advances in scalable architectures and hybrid methods may help broaden the adoption of data-driven scheduling in industrial settings.